\documentclass[letterpaper, 10 pt, conference]{ieeeconf}  
\pdfoutput=1

\usepackage{graphicx}
\usepackage{wrapfig}
\usepackage[percent]{overpic}
\usepackage{algorithm,algpseudocode,bm}

\newcommand{\pluseq}{\mathrel{+}=}

\IEEEoverridecommandlockouts                              

\usepackage{url}
\usepackage{breakurl}

\title{\LARGE \bf
Model Identification and Control of a Low-Cost Wheeled Mobile Robot Using Differentiable Physics
}

\author{Yanshi Luo and Abdeslam Boularias and Mridul Aanjaneya$^*$
\thanks{*Corresponding author}
\thanks{The authors are with the Department of Computer Science, Rutgers University, 110 Frelinghuysen Road, Piscataway, NJ 08854 USA
        {\small (email: yanshi.luo@rutgers.edu, ab1544@cs.rutgers.edu, mridul.aanjaneya@rutgers.edu). Fax: (732) 445 0537.}}%
}

\begin{document}

\maketitle
\thispagestyle{empty}
\pagestyle{empty}

\begin{abstract}

We present the design of a low-cost wheeled mobile robot, and an analytical model for predicting its motion under the influence of motor torques and friction forces. Using our proposed model, we show how to analytically compute the gradient of an appropriate loss function, that measures the deviation between predicted motion trajectories and real-world trajectories, which are estimated using Apriltags and an overhead camera. These analytical gradients allow us to automatically infer the unknown friction coefficients, by minimizing the loss function using gradient descent. Motion trajectories that are predicted by the optimized model are in excellent agreement with their real-world counterparts. Experiments show that our proposed approach is computationally superior to existing black-box system identification methods and other data-driven techniques, and also requires very few real-world samples for accurate trajectory prediction. The proposed approach combines
the data efficiency of analytical models based on first principles, with the flexibility of data-driven methods, which makes it appropriate for low-cost robots. 
Using the learned model and our gradient-based optimization approach, we show how to automatically compute motor control signals for driving the robot along pre-specified curves.


\end{abstract}


\section{Introduction}
\label{sec:introduction}

\begin{figure}
\includegraphics[width=\columnwidth]{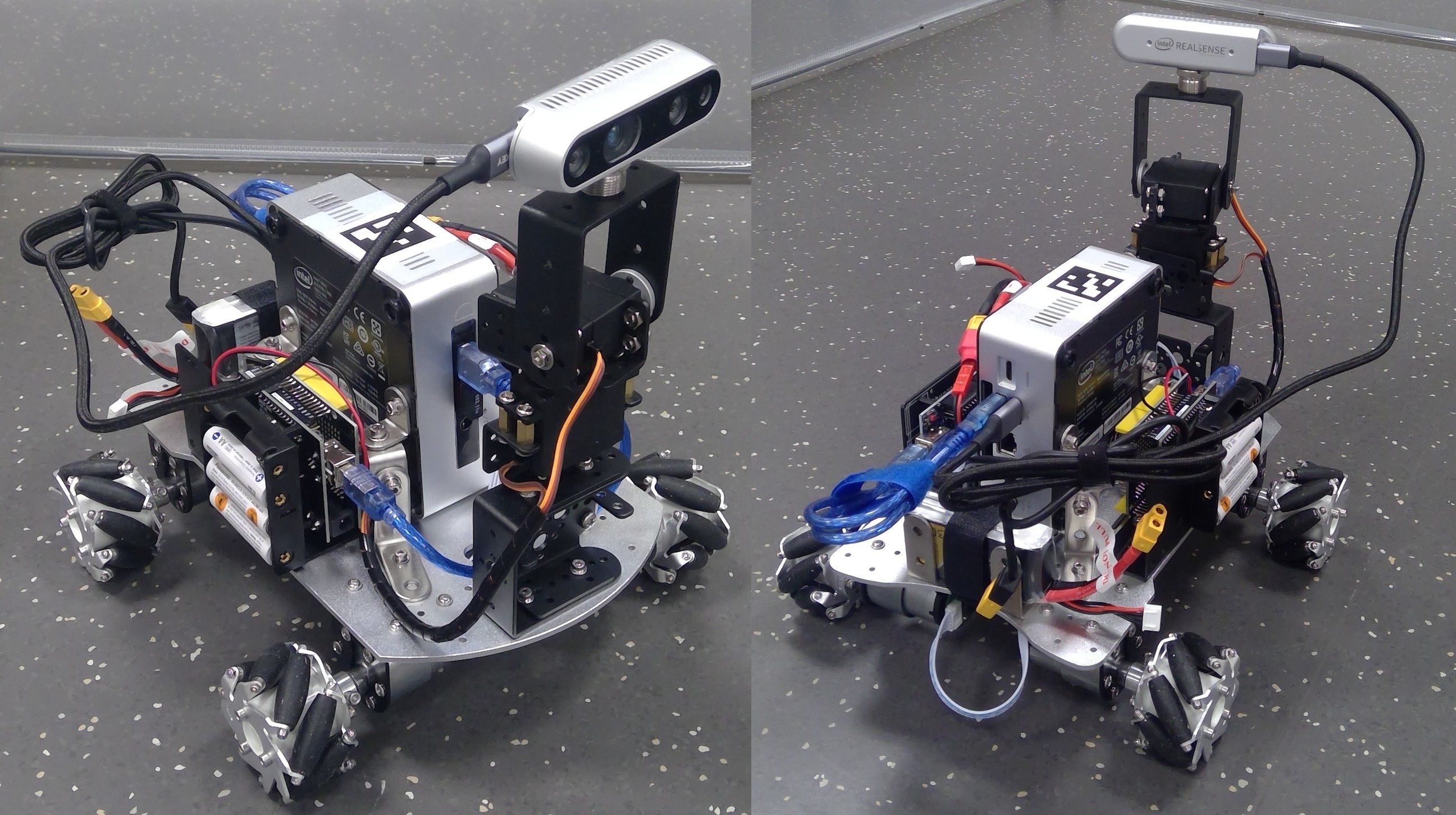}
\caption{\small Low-cost mobile robot, which costs approximately $\$1200$, designed and built in the present work, and used in the experiments.}
\label{fig:mobile-robot}
\end{figure}

With the availability of affordable 3D printers and micro-controllers such as Arduino~\cite{Banzi:2008:Arduino}, Beaglebone Black~\cite{Hamilton:2016:BBC}, and Raspberry Pi~\cite{Halfacree:2012:RPi}, light-weight high-performance computing platforms such as Intel's Next Unit of Computing (NUC)~\cite{NUC} and NVIDIA's Jetson Nano~\cite{Jetson-Nano}, and programmable RGB-D cameras, such as Intel's Realsense~\cite{Realsense}, there is renewed interest in building low-cost robots for various tasks~\cite{Adeept}. Motivated by these developments, the long-term goal of the present work is to develop affordable mobile robots that can be easily assembled using off-the-shelf components and 3D-printed parts. The assembled affordable robots will be used for exploration and scene understanding in unstructured environments. They can also be augmented with affordable robotic arms and hands for manipulating objects. Our ultimate objective is to remove the economic barrier to entry that has so far limited research in robotics to a relatively small number of groups that can afford expensive robot hardware. To that end, we have designed our own wheeled mobile robot for exploration and scene understanding, illustrated in Figure~\ref{fig:mobile-robot}. The first contribution of this work is then the hardware and software design of the proposed robot.

High-end robots can easily be controlled using software tools provided by manufacturers. Their physical properties, such as inertial and frictional parameters, are also precisely measured, which eliminates the need for further calibrations.
Affordable robots assembled and fabricated in-house are significantly more difficult to control due to uncertainties in the manufacturing process, which result in differences in size, weight and inertia according to the manufacturing technique. Due to this uncertainty, hand-crafting precise and shared models for these robots is challenging. For example, the wheels of the robot in Figure~\ref{fig:mobile-robot} cannot be precisely modeled manually because of their complex structure and uncertain material properties. 
Moreover, the frictions between the wheels and the terrain vary largely when the robot is deployed on an unknown non-uniform terrain. Statistical learning tools such as Gaussian processes and neural networks have been largely used in the literature to deal with this uncertainty and to learn dynamic and kinematic models directly from data. While such methods have the advantage of being less brittle than classical analytical models, they typically require large amounts of training data collected from each individual robot and for every type of terrain. 

In this work, we propose a hybrid data-driven approach that combines the versatility of machine learning techniques with the data-efficiency of physics models derived from first principles. The main component of the proposed approach is a self-tuning differentiable physics simulator of the designed robot. The proposed simulator takes as inputs the robot's pose and generalized velocity at a given time, a sequence of control signals, and returns a trajectory of predicted future poses and velocities. After executing the sequence of controls on the real robot, the resulting ground-truth trajectory of the robot is recorded and systematically compared
to the predicted one. The difference, known as the {\it reality gap}, between the predicted and the ground-truth trajectories is then used to automatically identify the unknown coefficients of friction between each of the robot's wheels and the present terrain. Since the identification process must happen on the fly and in real time, black-box optimization tools cannot be effectively used. Instead, we show how to compute analytically the derivatives of the reality gap with respect to each unknown coefficient of friction, and how to use these derivatives to identify the coefficients by following the gradient-descent technique. A key novelty of our approach is the integration of a differentiable forward kinematic model with a neural-network dynamic model. The kinematic model represents the part of the system that can be modeled analytically in a relatively easy manner, while the dynamics neural network is used for modeling the more complex relation between the frictions and the velocities. But since both parts of the system are differentiable, the gradient of the simulator's output is back-propagated all the way to the coefficients of friction and used to update them. 

The time and data efficiency of the proposed technique are demonstrated through two series of experiments that we have performed with the robot illustrated in Figure~\ref{fig:mobile-robot}. The first set of experiments consists in executing different control signals with the robot, and recording the resulting trajectories. Our technique is then used to identify the friction coefficients of each individual wheel, and to predict future trajectories accordingly. 
The second set of experiments consists in using the identified parameters in a model-predictive control loop to select control signals that allow the robot to track predefined trajectories.  The proposed gradient-based technique is shown to be more efficient computationally than black-box optimization methods, and more accurate than a neural network trained using the same small amount of data.

\section{Related Work}
\label{sec:related_work}

The problem of {\it learning dynamic and kinematic models of skid-steered robots} has been explored in several past works. Vehicle model identification by integrated prediction error minimization was proposed in~\cite{Seegmiller-2013-7754}. Rather than calibrate the system differential equation directly for unknown parameters, the approach proposed in~\cite{Seegmiller-2013-7754} calibrates its first integral. However, the dynamical model of the robot is approximated linearly using numerical first-order derivatives, in contrast with our approach that computes the gradient of the error function analytically. 
A relatively similar approach was used in~\cite{Seegmiller-2014-7905} for calibrating a kinematic wheel-ground contact model for slip prediction. 
The present work builds on the kinematic model for feedback control of an omnidirectional wheeled mobile robot proposed in~\cite{Muir1987KinematicMF}.

A learning-based {\it Model Predictive Control} (MPC) was used in~\cite{SchoelligJFR2015} to control a mobile robot in challenging outdoor environments. The proposed model uses a simple a priori vehicle model and a learned disturbance model, which is modeled as Gaussian Process and learned from local data. An MPC technique was also applied to the autonomous racing problem in simulation in~\cite{RosoliaCB16}. The proposed system identification technique consists in decomposing the dynamics as the sum of a known deterministic function and noisy residual that is learned from data by using the least mean square technique. 
The approach proposed in the present work shares some similarities with the approach presented in~\cite{Xieetal_ICRA2016}, wherein the dynamics equations of motion are used to analytically compute the mass of a robotic manipulator. A neural network was also used in a prior work to calibrate the wheel–terrain interaction frictional term of skid-steered dynamic model \cite{ORDONEZ2017207}. An online estimation method that identifies key
terrain parameters using on-board robot sensors is presented in~\cite{1339393}. A simplified form of classical terramechanics equations was used along with a linear-least squares method for terrain parameters estimation. Unlike in the present work that considers a full body simulation, \cite{1339393} considered only a model of a rigid wheel on deformable terrains. A dynamic model is also presented in~\cite{1019459} for omnidirectional wheeled mobile robots, including surface slip. However, the friction coefficients in~\cite{1019459} were experimentally measured, unlike in the present work where the friction terms are automatically tuned from data by using the gradient of the distance between simulated trajectories and the observed ones. 

\emph{Classical system identification} builds a dynamics model by
minimizing the difference between the model's output signals and real-world response data for the same input signals~\cite{swevers1997optimal,Ljung:1999:SIT:293154}. Parametric rigid body dynamics models have also been combined with non-parametric model learning for approximating the inverse dynamics of a robot~\cite{nguyen2010using}. 

There has been a recent surge of interest in developing \emph{natively differentiable physics engines}. For example,~\cite{DegraveHDW16} used the Theano framework to develop a physics engine that can be used to differentiate control parameters in robotics applications. The same framework  can be altered to differentiate model parameters instead.
This engine is implemented for both CPU and GPU, and it has been shown  how such an engine speeds up the optimization process for finding optimal policies. 
A combination of a learned and a differentiable simulator was used to predict action effects on planar objects~\cite{DBLP:journals/corr/abs-1710-04102}. 
A differentiable framework for integrating fluid dynamics with deep networks was also used to learn fluid parameters from
data, perform liquid control tasks, and learn policies to manipulate liquids~\cite{Schenck2018SPNetsDF}.
Differentiable physics simulations were also used for manipulation planning and tool use~\cite{18-toussaint-RSS}. 
Recently, it has been observed that a standard physical simulation formulated as a Linear Complementary Problem (LCP) is also differentiable and can be implemented in PyTorch~\cite{Belbute-Peres2017}. In~\cite{Mordatch:2012}, a differentiable contact model was used to allow for optimization of several  locomotion and manipulation tasks. The present work is another step toward the adoption of self-tuning differentiable physics engines as both a data-efficient and time-efficient tool for learning and control in robotics.

\section{Robot Design}
\label{sec:design}

The design of our mobile robot consists of the hardware assembly, and the software drivers that provide control signals for actuation. These components are described below.

\begin{figure}[!h]
\includegraphics[width=\columnwidth]{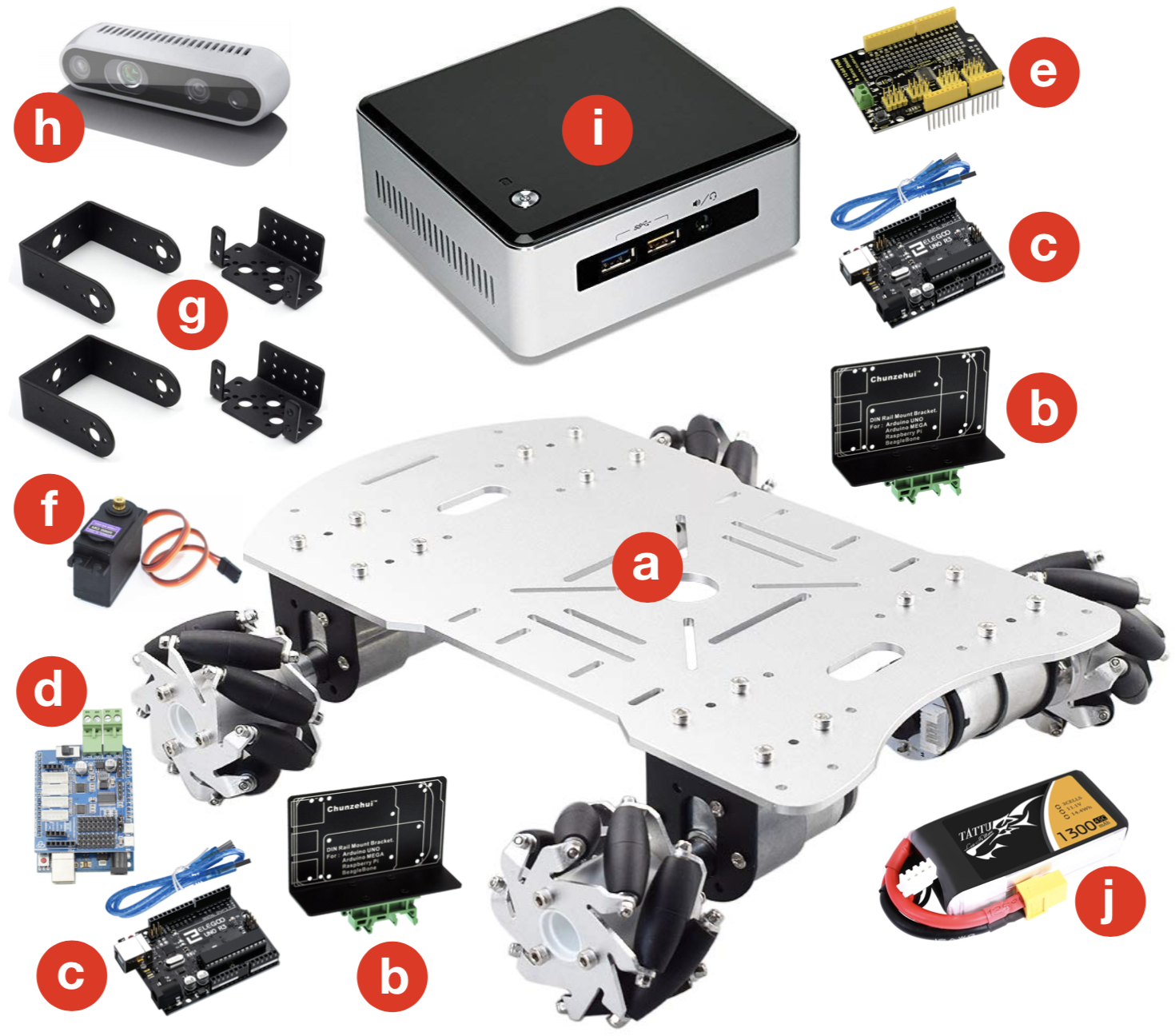}
\caption{\small Various hardware components for the mobile robot. All hardware and electronic designs were performed in-house at Rutgers University by the co-authors of the present paper.}
\label{fig:hardware-assembly}
\end{figure}
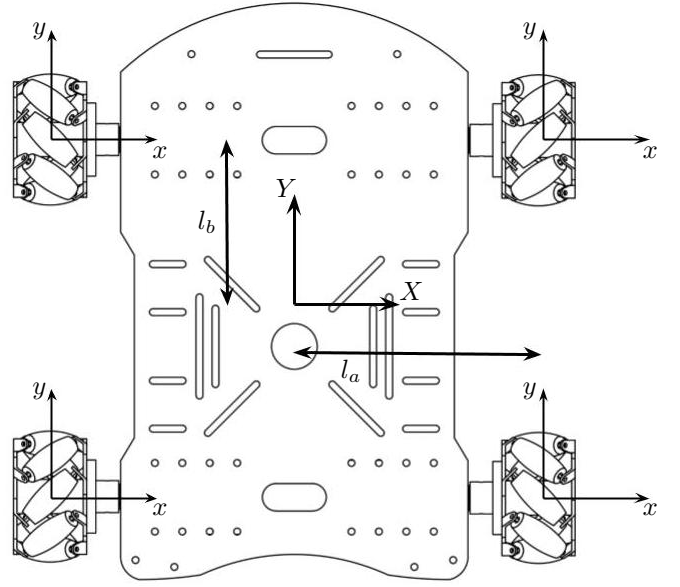
\begin{figure}
\begin{overpic}[width=\columnwidth]{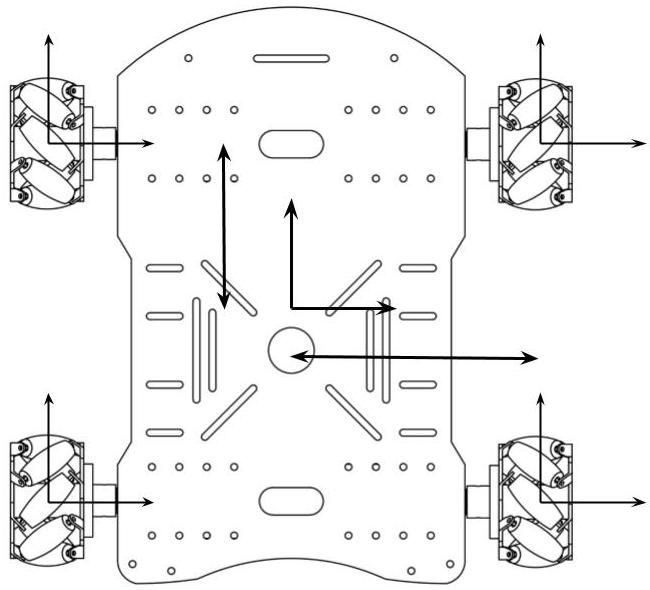}
    \put(30,55){$l_b$}
    \put(52,32){$l_a$}
    \put(61,44){$X$}
    \put(42,60){$Y$}
    \put(80,85){$y$}
    \put(98.5,66){$x$}
    \put(80,29.5){$y$}
    \put(98.5,11){$x$}
    \put(4.5,29.5){$y$}
    \put(23,11){$x$}
    \put(4.5,85){$y$}
    \put(23,66){$x$}
\end{overpic}
\caption{\small Top view of the mobile robot.}
\label{fig:chassis}
\end{figure}

\subsection{Hardware Assembly}

The robot is built from various components, as shown in Figure~\ref{fig:hardware-assembly}, which comprise: (a) a central chassis with four Mecanum omni-directional wheels that are run by $12V$ DC motors, (b) two rail mount brackets for Arduino, (c) two Arduino UNOs, (d) Arduino shield for the DC motors, (e) servo motor drive shield for Arduino, (f) two MG996R servos, (g) two 2DOF servo mount brackets, (h) Intel Realsense D435, (i) Intel NUC5i7RYH, and (j) two $1300$mAh, $11.1V$ DC batteries. One battery powers the DC motors, while the other one powers the Intel NUC. The servo motors are powered separately with a $6V$ DC source. All these components can be purchased from different vendors through Amazon.

\subsection{Control Software}

The two Arduinos that drive the servos and the DC motors are connected to the Intel NUC via USB cables. To send actuation commands from the NUC to the Arduinos, we use the single byte writing Arduino method \texttt{Serial.write()}, following the custom Serial protocol developed in~\cite{ARR}. Each message is encoded as one byte, and the corresponding  message parameters are then sent byte per byte, which are reconstructed upon reception using bitwise shift and mask operations. The limited buffer size of the Arduino is also accounted for, by having the Arduino ``acknowledge'' receipt of each message. We have generalized the implementation in~\cite{ARR} to support two Arduinos, two servo motors, and the differential drive mechanism for the wheels.
\section{Analytical Model}
\label{sec:model}

\begin{wrapfigure}[13]{r}{0.52\columnwidth} 
\vspace*{-4mm}
\begin{overpic}[width=0.52\columnwidth]{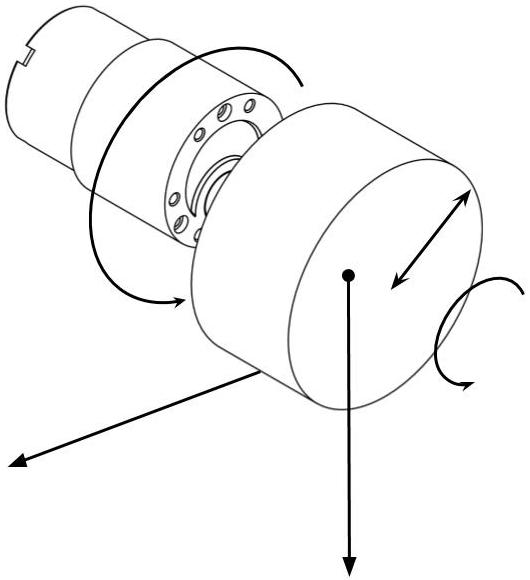}
    \put(30,95){$T = T_s\left(1 - \frac{\omega}{\omega_s}\right)$}
    \put(62,15){$\frac{Mg}{4}$}
    \put(8,18){$\mu N = \mu\frac{Mg}{4}$}
    \put(67.5,60){$R$}
    \put(80,37){$\omega$}
\end{overpic}
\vspace*{-7mm}
\caption{\small Motor torque and friction on a simplified cylindrical wheel.}
\label{fig:wheel}
\end{wrapfigure}
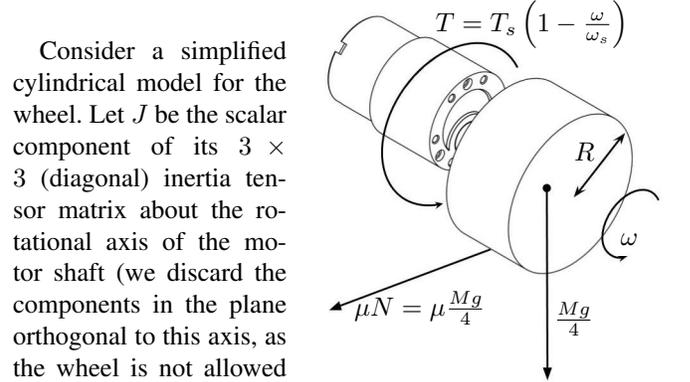
Consider a simplified cylindrical model for the wheel. Let $J$ be the scalar  component of its $3\times 3$ (diagonal) inertia tensor matrix about the rotational axis of the motor shaft (we discard the components in the plane orthogonal to this axis, as the wheel is not allowed to rotate in this plane). Then, its equation of motion can be written as:

\begin{equation}
\label{eqn:motor-pde}
J\frac{d\omega}{dt} = T_s\left(1 - \frac{\omega}{\omega_s}\right) - \mu \frac{Mg}{4}R
\end{equation}
where $T_s$ is the motor stall torque, $\omega$ is the wheel's angular velocity, $\omega_s$ is the desired angular velocity (specified by the PWM signal), $\mu$ is the coefficient of friction, $g$ is the acceleration due to gravity, $R$ is the wheel radius, and $M$ is the mass of the robot. We assume that the weight of the robot is balanced uniformly by all four wheels (thus, the term $Mg/4$). The first term on the right hand side was derived in~\cite{Rojas:DC}. Equation (\ref{eqn:motor-pde}) can be analytically integrated to obtain:

\begin{equation}
\label{eqn:omega-expr-transient}
\omega = \omega_s\left(1 - \frac{\mu M g R}{4 T_s}\right)\left(1 - \exp\left(-\frac{T_st}{J\omega_s}\right)\right)
\end{equation}
Since we are interested in large time spans for mobile robot navigation, only the steady state terms in equation (\ref{eqn:omega-expr-transient}) are important. Thus, we discard the transient terms to arrive at:

\begin{equation}
\label{eqn:omega-expr-steady-state}
\omega = \omega_s\left(1 - \frac{\mu M g R}{4 T_s}\right)
\end{equation}
which gives us an expression for the wheel's angular velocity as a function of the ground friction forces. We assume that the friction coefficient $\mu_j$ for each wheel $j$ is different. The chassis of our mobile robot is similar to that of the \emph{Uranus} robot, that was developed in the Robotics Institute at Carnegie Mellon University~\cite{Muir1987KinematicMF}. The following expression was derived in~\cite{Muir1987KinematicMF} for the linear and angular velocity of the Uranus robot, using the wheel angular velocities as input:

\begin{equation}
\label{eqn:kinematic-cmu}
\left[
\begin{array}{c}
     v_x \\
     v_y \\
     \omega_z
\end{array}
\right] = \underbrace{\frac{R}{4l_{ab}}\left[\begin{array}{cccc}
-l_{ab} & l_{ab} & -l_{ab} & l_{ab} \\
l_{ab} & l_{ab} & l_{ab} & l_{ab} \\
1 & -1 & -1 & 1
\end{array}
\right]}_B\left[\begin{array}{c}
\omega_1 \\
\omega_2 \\
\omega_3 \\
\omega_4
\end{array}
\right]
\end{equation}
where $l_{ab} = l_a + l_b$ and $l_a, l_b$ are as defined in Figure~\ref{fig:chassis}, $v_x,v_y$ are the linear velocity components of the robot along the $X$ and $Y$ axes, and $\omega_z$ is the angular velocity of the robot about the $Z$ axis. Equations (\ref{eqn:omega-expr-steady-state}), (\ref{eqn:kinematic-cmu}) define a motion model for our mobile robot including the effects of friction.
\section{Loss Function}
\label{sec:loss}

Let $(x^i,y^i,\theta^i)$ be the generalized position of the mobile robot at time $t^i$, and $(v_x^i,v_y^i,\omega_z^i)$ be its generalized velocity. Then, its predicted state at time $t^{i+1}$ can be computed as:

\begin{equation}
\label{eqn:model-prediction}
\left[
\setlength{\tabcolsep}{3pt}
\begin{tabular}{c}
$x^{i+1}$ \\
$y^{i+1}$ \\
$\theta^{i+1}$
\end{tabular}
\right] = \left[
\setlength{\tabcolsep}{3pt}
\begin{tabular}{c}
$x^i$ \\
$y^i$ \\
$\theta^i$
\end{tabular}
\right] + \Delta t\left[
\setlength{\tabcolsep}{3pt}
\begin{tabular}{ccc}
$\cos\theta^i$ & -$\sin\theta^i$ & 0 \\
$\sin\theta^i$ & $\cos\theta^i$ & 0 \\
0 & 0 & 1
\end{tabular}
\right]\left[
\setlength{\tabcolsep}{3pt}
\begin{tabular}{c}
$v_x^i$ \\
$v_y^i$ \\
$\omega_z^i$
\end{tabular}
\right]
\end{equation}
where $\Delta t = t^{i+1} - t^i$. The loss function computes the {\it simulation-reality gap}, defined as the divergence of a predicted robot's trajectory from an observed ground-truth one. The loss is defined as follows,

\begin{equation}
\label{eqn:loss-gt}
L_{g t}=\sum_{k=1}^{N}\left(\left|x^{k}-x_{g t}^{k}\right|^{2}+\left|y^{k}-y_{g t}^{k}\right|^{2}\right)^{1 / 2}
\end{equation}
where $(x^k_{gt},y^k_{gt})$ are the ground-truth position values at time-step $k$, estimated using Apriltags~\cite{wang2016iros} and an overhead camera. $(x^k,y^k)$ are the positions predicted in simulation by using Equation~\ref{eqn:model-prediction} and the same sequence of control signals as the ones provided to the real robot. The video is recorded using the overhead camera at $60$fps, whereas the PWM signal is sent to the motor at a 3-4$\times$ higher frequency. Thus, there are several simulation time steps between two consecutive ground truth position values.

In practice, we observed that the predicted state can often ``lag behind'' the ground truth values, even when the robot is \emph{exactly} following the overall path, leading to a high loss value. Thus, we fit a spline curve to the ground truth values and compute another loss function that uses the point $(x^k_{sp},y^k_{sp})$ \emph{closest} to this curve from the predicted state at time-step $k$:

\begin{equation}
\label{eqn:loss-spline}
L_{s p}=\sum_{k=1}^{N}\left(\left|x^{k}-x_{s p}^{k}\right|^{2}+\left|y^{k}-y_{s p}^{k}\right|^{2}\right)^{1 / 2}
\end{equation}
Note that the loss $L_{sp}$ alone is not sufficient, as it does not penalize the simulated robot for not moving at all from its starting position. Thus, we use a weighted linear combination of $L_{gt}$ and $L_{sp}$ as our actual loss function, as given below:

\begin{equation}
\label{eqn:loss}
L=w_{1} \cdot L_{s p}+w_{2} \cdot L_{g t}
\end{equation}
where we set $w_1=0.8, w_2=0.2, M=4, g=9.8, r=0.03$, and $T_s=0.6$. The pseudocode for loss computation is shown in Algorithm~\ref{alg:loss-computation}. Vector quantities are shown in bold.
$B$ is defined in Equation~\ref{eqn:kinematic-cmu}, and $\mathbf{\mu}$ is a vector of the friction coefficients of the different wheels. 
Line 4 uses component-wise vector multiplication. $T$ is the total number of time-steps, and $t^i$ is the time when the PWM signal was applied to the motor at time-step $i$. The function \texttt{IsGroundTruthSample} checks if a ground truth Apriltag estimate exists at time $t^i$, and if so, then the function \texttt{GroundTruthIndex} returns the index of that estimate.

\begin{algorithm}
\caption{\texttt{LossComputation}($B,\boldsymbol{\mu}$)}
\label{alg:loss-computation}
\begin{algorithmic}[1]
\State Initialize $l\leftarrow 0$, $p\leftarrow (x^0_{gt},y^0_{gt},\theta^0_{gt})$
\For{$i=1\ldots T$}
\State Compute $\Delta t\leftarrow t^{i+1} - t^i$
\State Compute $\boldsymbol{\omega}\leftarrow\boldsymbol{\omega}_s\left(\mathbf{1} - \boldsymbol{\mu}Mgr/4T_s\right)$
\State Compute $\left(v_{x}, v_{y}, \omega_{z}\right) \leftarrow B \boldsymbol{\omega}$
\State $s\leftarrow\sin(p_2),c\leftarrow\cos(p_2)$
\If{\texttt{IsGroundTruthSample}($i$)}
\State $k\leftarrow \texttt{GroundTruthIndex}(i)$
\State $(\Delta p_x,\Delta p_y )\leftarrow (p_0 - x^k_{gt},p_1 - y^k_{gt})$
\State $(\Delta q_x,\Delta q_y )\leftarrow (p_0 - x^k_{sp},p_1 - y^k_{sp})$
\State $l \pluseq w_1\left(\Delta q_x^2 + \Delta q_y^2\right)^{1/2} + w_2\left(\Delta p_x^2 + \Delta p_y^2\right)^{1/2}$
\EndIf
\State $p\pluseq \Delta t(c v_x - s v_y,s v_x + c v_y,\omega_z)$
\EndFor \\
\Return $l$
\end{algorithmic}
\end{algorithm}
\section{Differentiable Physics}
\label{sec:diff_physics}

To minimize the loss function in equation (\ref{eqn:loss}) with respect to unknown parameters $\mu_j$, corresponding to the friction values for the four wheels, we derive analytical expressions for the \emph{gradient} of the loss with respect to each variable $\mu_j$:

\begin{equation}
\label{eqn:loss-gradient}
\frac{\partial L}{\partial\mu_j} = w_1\cdot\frac{\partial L_{sp}}{\partial\mu_j} + w_2\cdot\frac{\partial L_{gt}}{\partial\mu_j}
\end{equation}
Let $(\Delta x^k_{gt},\Delta y^k_{gt}) = (x^k - x^k_{gt},y^k - y^k_{gt})$ and $d^k$ denote the length $\left(\vert \Delta x^k\vert^2 + \vert \Delta y^k\vert^2\right)^{1/2}$. Then the second term in equation (\ref{eqn:loss-gradient}) can be expanded using the chain rule as follows:

\begin{equation}
\label{eqn:loss-gt-gradient}
\frac{\partial L_{gt}}{\partial\mu_j} = \sum_{k=1}^N \frac{1}{d^k}\left(\Delta x^k_{gt}\cdot\frac{\partial x^k}{\partial\mu_j} + \Delta y^k_{gt}\cdot\frac{\partial y^k}{\partial\mu_j}\right)
\end{equation}
The derivatives in equation (\ref{eqn:loss-gt-gradient}) can be computed using equations (\ref{eqn:kinematic-cmu}), (\ref{eqn:model-prediction}) at the higher frequency of the PWM signal sent to the motors, used for predicting the next state, as:

\begin{eqnarray}
\label{eqn:gradient-x}
\frac{\partial x^{i+1}}{\partial \mu_j} &=& \frac{\partial x^i}{\partial \mu_j} + \Delta t\cos\theta^i b_{1j}\frac{\partial\omega^i}{\partial\mu_j} - \Delta t\sin\theta^i b_{2j}\frac{\partial\omega^i}{\partial\mu_j} \nonumber \\
 &-& \Delta t\left(v_x^i\sin\theta^i + v_y^i\cos\theta^i\right)\frac{\partial\theta^i}{\partial\mu_j} \\
\label{eqn:gradient-y}
\frac{\partial y^{i+1}}{\partial\mu_j} &=& \frac{\partial y^i}{\partial\mu_j} + \Delta t\sin\theta^i b_{1j}\frac{\partial\omega^i}{\partial\mu_j} + \Delta t\cos\theta^i b_{2j}\frac{\partial\omega^i}{\partial\mu_j} \nonumber \\
&+& \Delta t\left(v_x^i\cos\theta^i - v_y^i\sin\theta^i\right)\frac{\partial\theta^i}{\partial\mu_j} \\
\label{eqn:gradient-theta}
\frac{\partial\theta^{i+1}}{\partial\mu_j} &=& \frac{\partial\theta^i}{\partial\mu_j} + \Delta t b_{3j}\frac{\partial\omega^i}{\partial\mu_j}
\end{eqnarray}
where $b_{rs}$ is the $(r,s)$ entry in the matrix $B$, as defined in equation (\ref{eqn:kinematic-cmu}). The derivative $\partial\omega^i/\partial\mu_j$ can be computed using equation (\ref{eqn:omega-expr-steady-state}). The expression for $\partial L_{sp}/\partial\mu_j$ in equation (\ref{eqn:loss-gradient}) can be derived similarly. Note that, strictly speaking, the closest point $(x^k_{sp},y^k_{sp})$ on the spline curve is a \emph{function} of the point $(x^k,y^k)$. However, we have empirically found that estimating its derivative is not necessary and can be ignored, when computing the term $\partial L_{sp}/\partial\mu_j$, for $j\in\{1\ldots4\}$.

\begin{algorithm}
\caption{\texttt{GradientComputation}($B,\boldsymbol{\mu}$)}
\label{alg:gradient-computation}
\begin{algorithmic}[1]
\State Initialize $\boldsymbol{g}\leftarrow [0]_{4\times1}$, $J\leftarrow [0]_{3\times4}$, $p\leftarrow (x^0_{gt},y^0_{gt},\theta^0_{gt})$
\For{$i=1\ldots T$}
\State Compute $\Delta t\leftarrow t^{i+1} - t^i$
\State Compute $\boldsymbol{\omega}\leftarrow\boldsymbol{\omega}_s\left(1 - \boldsymbol{\mu}Mgr/4T_s\right)$
\State Compute $\left(v_{x}, v_{y}, \omega_{z}\right) \leftarrow B \boldsymbol{\omega}$
\State Compute $\boldsymbol{d\omega}\leftarrow - \boldsymbol{\omega}_sMgr/4T_s$
\State $s\leftarrow\sin(p_2),c\leftarrow\cos(p_2)$
\For{$j=0\ldots3$}
\State $J_{0j} \pluseq \Delta t\left\{ \boldsymbol{d\omega}_j(c B_{0j} - s B_{1j}) - (v_x s +v_y c)J_{2j}\right\}$
\State $J_{1j} \pluseq \Delta t\left\{ \boldsymbol{d\omega}_j(s B_{0j} + c B_{1j}) + (v_x c +v_y s)J_{2j}\right\}$
\State $J_{2j} \pluseq \Delta t B_{2j}\boldsymbol{d\omega}_j$
\EndFor
\If{\texttt{IsGroundTruthSample}($i$)}
\State $k\leftarrow \texttt{GroundTruthIndex}(i)$
\State $(\Delta p_x,\Delta p_y )\leftarrow (p_0 - x^k_{gt},p_1 - y^k_{gt})$
\State $(\Delta q_x,\Delta q_y )\leftarrow (p_0 - x^k_{sp},p_1 - y^k_{sp})$
\State $d_{gt} \leftarrow \left(\Delta p_x^2 + \Delta p_y^2\right)^{1/2}$
\State $d_{sp} \leftarrow \left(\Delta q_x^2 + \Delta q_y^2\right)^{1/2}$
\vspace{1mm}
\State $\boldsymbol{g} \pluseq J[\{0,1\},:]^T\left(\frac{w_1}{d_{sp}}\left[
\setlength{\tabcolsep}{1pt}
\begin{tabular}{c}
$\Delta q_x$ \\
$\Delta q_y$
\end{tabular}\right] + \frac{w_2}{d_{gt}}\left[
\setlength{\tabcolsep}{1pt}
\begin{tabular}{c}
$\Delta p_x$ \\
$\Delta p_y$
\end{tabular}\right]\right)$
\EndIf
\State $p\pluseq \Delta t(c v_x - s v_y,s v_x + c v_y,\omega_z)$
\EndFor \\
\Return $\boldsymbol{g}$
\end{algorithmic}
\end{algorithm}

\begin{figure*}[!t]
\setlength{\tabcolsep}{1pt}
\begin{tabular}{cccc}
\includegraphics[width=.49\columnwidth]{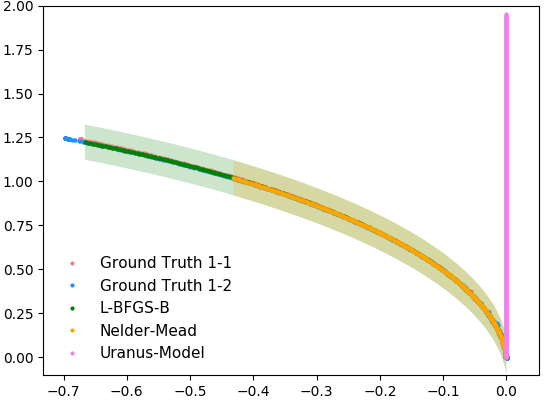} & \includegraphics[width=.49\columnwidth]{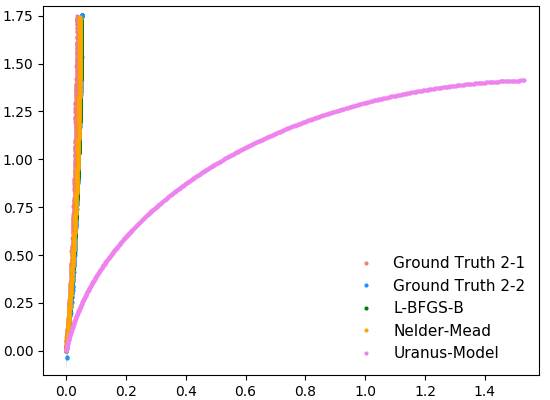} &
\includegraphics[width=.49\columnwidth]{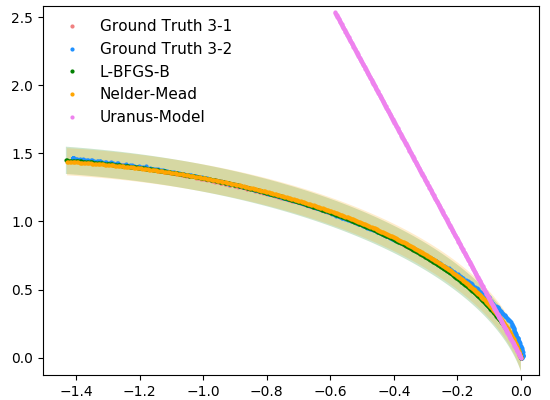} &
\includegraphics[width=.49\columnwidth]{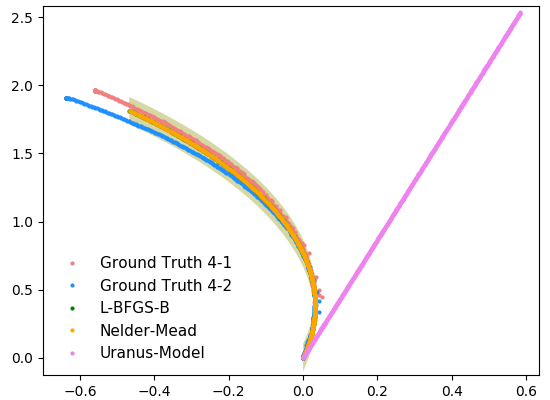} \\
(a) & (b) & (c) & (d) \\
\includegraphics[width=.49\columnwidth]{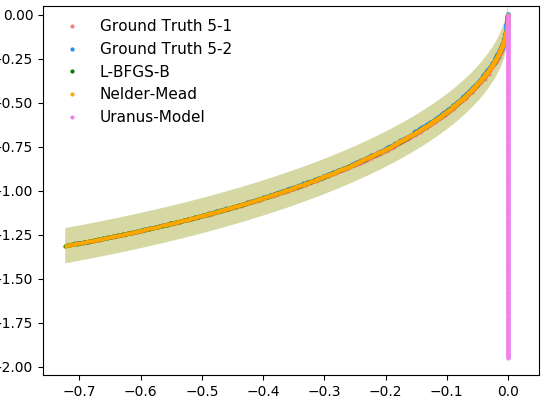} &
\includegraphics[width=.49\columnwidth]{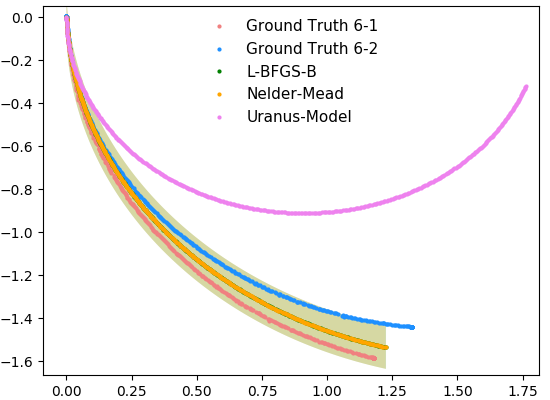} &
\includegraphics[width=.49\columnwidth]{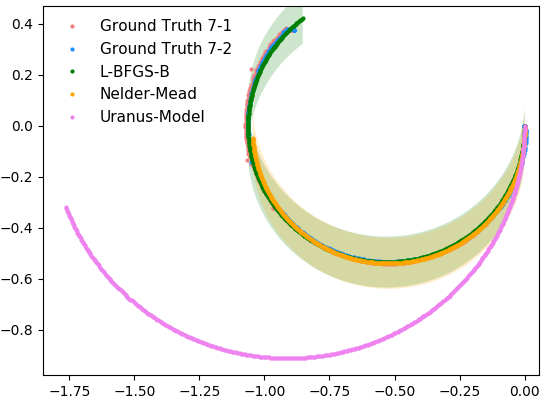} &
\includegraphics[width=.49\columnwidth]{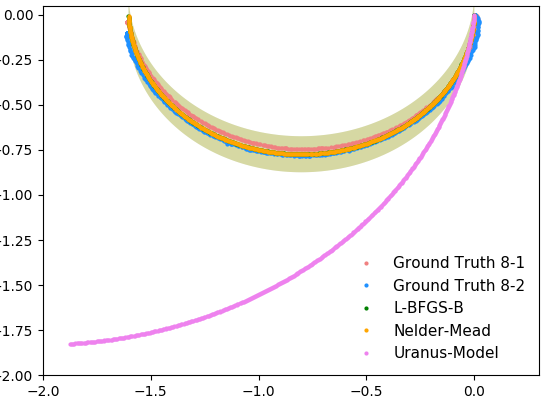} \\
(e) & (f) & (g) & (h)
\end{tabular}
\vspace*{-2mm}
\caption{\small Robot trajectories for 8 different motor control signals. Ground truth values estimated using Apriltags are shown in red and blue. Our method uses L-BFGS-B to estimate friction parameters, where we compute the gradient using Algorithm~\ref{alg:gradient-computation}; the result is shown in green. For comparison, we also show the result of the Nelder-Mead method in yellow, which is a derivative-free optimization technique. Trajectories predicted by L-BFGS-B and the Nelder-Mead method are overlapping for the plots shown in (b), (d), (e), (f) and (h).}
\label{fig:ground-truth}
\vspace*{-2mm}
\end{figure*}

\begin{figure}
\includegraphics[width=\columnwidth]{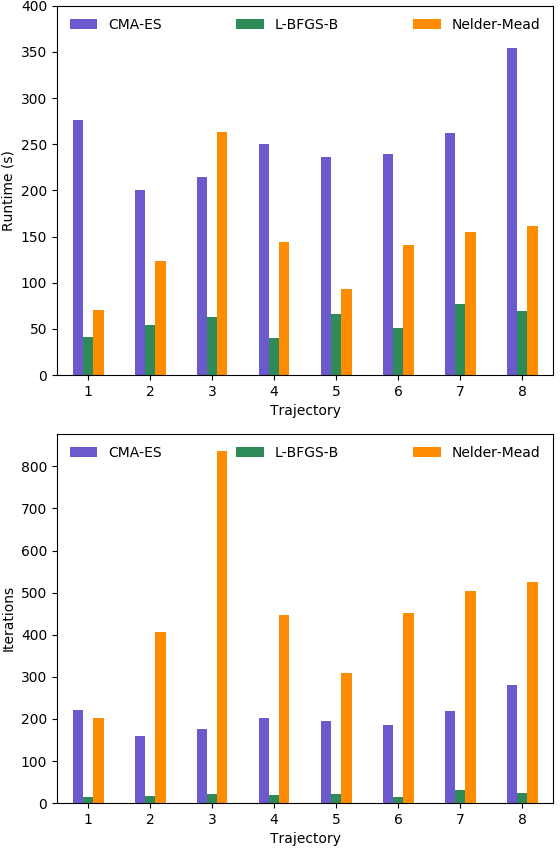}
\vspace*{-7mm}
\caption{\small Run-time (in seconds) and total iteration counts for estimating friction coefficients for all $8$ trajectories using CMA-ES, L-BFGS-B with our gradient function, and the Nelder-Mead method. Note that all three methods use our analytical model, but only L-BFGS-B makes use of our analytical gradients.}
\label{fig:runtime-and-iteration}
\end{figure}
\begin{figure}
\includegraphics[width=\columnwidth]{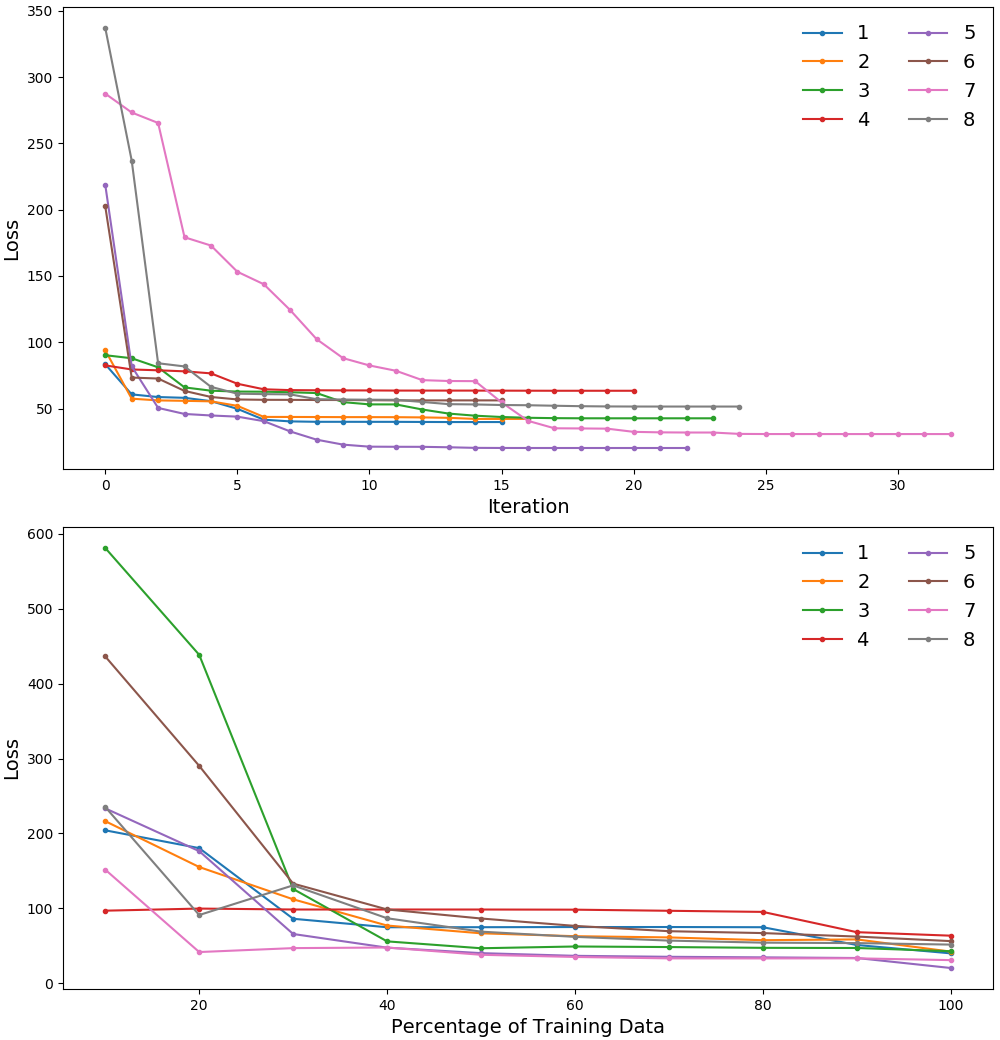}
\caption{\small (Top) Loss value vs iteration for our method, which makes rapid progress in the initial few iterations. (Bottom) Efficiency of our method with less training data, according to the percentage shown on the X-axis. Our method is potentially applicable in real-time settings for dynamically detecting changes in friction.}
\label{fig:LBFGS-loss}
\end{figure}

As shown in Section~\ref{sec:results}, using the gradients derived in Section~\ref{sec:diff_physics} to minimize the loss function in equation (\ref{eqn:loss}) yields friction parameters that give good agreements with the experimental trajectory that is estimated using Apriltags. However, we observed that the computed friction parameters can differ in values for two different trajectories (with different control signals). This implies that the friction $\mu_j$ for each wheel is not a constant, but a function $\mu_j(\omega_s)$ of the applied control signal. Thus, we first generate a sequence of trajectories with fixed control signals, and estimate friction parameters for each of them by separately minimizing the loss function in equation (\ref{eqn:loss}) using gradient descent. We then train a small neural network with 4 input nodes, 16 hidden nodes, and 4 output nodes. The input to the neural network are the applied control signals to the wheels, and the output are the friction parameters estimated using gradient descent. As shown in Section~\ref{sec:results}, a sequence of only $8$ input trajectories is enough to obtain reasonable predictions from the neural network, and allowed us to autonomously drive the mobile robot along an ``eight curve'' with high precision.

The pseudocode for gradient computation is shown in Algorithm~\ref{alg:gradient-computation}. Matrix $J$ stores the accumulated gradients for the generalized position, computed in equations (\ref{eqn:gradient-x})-(\ref{eqn:gradient-theta}). The notation $J[\{0,1\},:]$ refers to the first two rows of $J$.

\section{Experimental Results}
\label{sec:results}

We remotely controlled the robot for 6-8 seconds and collected 8 different trajectories, with 2 samples for each trajectory that were estimated using Apriltags~\cite{wang2016iros}, as shown in Figure~\ref{fig:ground-truth} in red and blue. The control signals applied to the wheels were constant for the entirety of each trajectory, but different per trajectory.  This was done for simplicity, as changing the control signals changes the direction of motion, and the robot may have driven out of the field of view.
A video of these experiments is attached to the present paper as a supplementary material.
\begin{figure}
\includegraphics[width=\columnwidth]{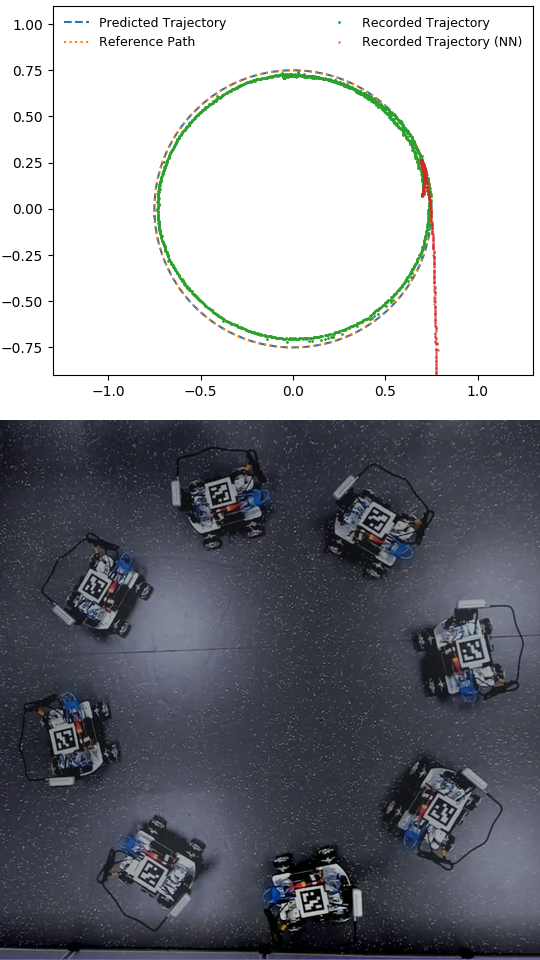}
\caption{\small (Top) Reference curve (orange), trajectory predicted using the proposed method (blue), real-world recorded trajectory using control signals computed by the proposed method (green), and a pure data-driven approach (red). (Bottom) Snapshots of the robot autonomously following a circular path using the proposed method.}
\label{fig:path-following-circle}
\end{figure}
\begin{figure}
\includegraphics[width=\columnwidth]{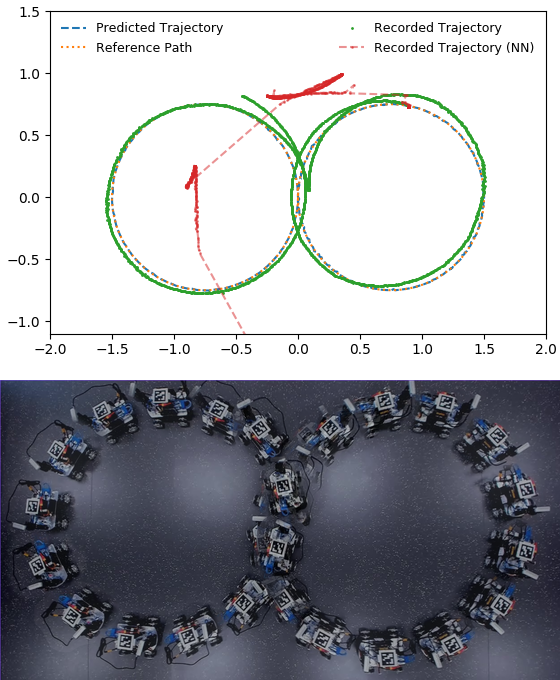}
\caption{\small (Top) Reference path (orange), trajectory predicted by the proposed method (blue), real-world recorded trajectory using control signals computed by the proposed method (green), and a pure data-driven approach (red). (Bottom) Snapshots of the robot autonomously following a ``figure 8'' using the proposed method.}
\label{fig:path-following-eight}
\end{figure}
\subsection{Model Identification}

To estimate the unknown friction coefficients, we use the L-BFGS-B optimization method, where the gradient is computed using Algorithm~\ref{alg:gradient-computation}. The result of our method is shown in Figure~\ref{fig:ground-truth} in green. We additionally imposed the constraints that all estimated friction values should lie in the interval $[0,2]$. Such constraints are naturally supported by L-BFGS-B. Our chosen values for $M, g, r, T_s$ ensure that $\omega_j\rightarrow0$ as $\mu_j\rightarrow2$ for $j\in\{1\ldots4\}$. For comparison, we show the result of the Nelder-Mead method in yellow, which is an unconstrained derivative-free optimization technique. To ensure that our problem is still well-defined in an unconstrained setting, we slightly modified equation (\ref{eqn:omega-expr-steady-state}) as:

\begin{equation}
\label{eqn:omega-expr-sigmoid}
\omega = \omega_s\left(1 - \frac{\sigma(\mu) M g R}{2 T_s}\right)
\end{equation}
where $\sigma(\mu)$ is the \emph{sigmoid} function, which always lies in the interval $[0,1]$ for all values of $\mu\in(-\infty,\infty)$. We also show the trajectories computed from the original Uranus motion model~\cite{Muir1987KinematicMF}, defined by equation (\ref{eqn:kinematic-cmu}). Since it does not account for friction, its predictions significantly deviate from the observed trajectories. The total run-time and iteration counts for our method, Nelder-Mead, and CMA-ES~\cite{Hansen:2003:CMA-ES}, which is also a derivative-free optimization method, are highlighted in Figure~\ref{fig:runtime-and-iteration}. As shown, our method requires very few iterations to converge, and is generally faster than both Nelder-Mead, which sometimes fails to converge (in 3 out of our 8 cases), and CMA-ES.
We did not show the trajectories predicted using CMA-ES in Figure~\ref{fig:ground-truth} to avoid clutter, as it converges to the same answer as L-BFGS-B, just takes longer, as shown in Figure~\ref{fig:runtime-and-iteration}.
Figure~\ref{fig:LBFGS-loss}(top) shows the loss value with increasing iteration counts of L-BFGS-B. The use of accurate analytic gradients allows for rapid progress in the initial few iterations itself. Figure~\ref{fig:LBFGS-loss}(bottom) shows the effect on loss when the training data is reduced according to the percentage on the X-axis. As shown, our method converges to almost the final loss value with only $40\%$ of the total data, making it \emph{data efficient}, and potentially applicable in real-time settings for dynamically detecting changes in the friction of the terrain.

\subsection{Path Following}

We also used our learned model to compute control signals, such that the robot could autonomously follow pre-specified curves within a given time budget $T$. We take as input the number of way-points $n$ that the robot should pass in a second, and discretize the given curve with $nT$ way-points. We assume that the control signal is constant between consecutive way-points. To compute the control signals, we again use L-BFGS-B, but modify Algorithms~\ref{alg:loss-computation} and~\ref{alg:gradient-computation} to optimize for the control signals $\boldsymbol{\omega}_s$, instead of the friction coefficients $\boldsymbol{\mu}$. Apart from changing the primary variable from $\boldsymbol{\mu}$ to $\boldsymbol{\omega}_s$, the only other change required is to use the derivative of equation (\ref{eqn:omega-expr-steady-state}) with respect to $\boldsymbol{\omega}_s$ in Algorithm~\ref{alg:gradient-computation}.

Figures~\ref{fig:path-following-circle} and~\ref{fig:path-following-eight} illustrate our results when the specified path is a circle, and a more challenging ``figure 8''. Shown are the reference path, the path predicted by using the control signals computed from our learned model, and the real robot trajectory estimated with Apriltags after applying these control signals. Our learned model is accurate enough that the robot can successfully follow the specified path very closely. To test the robustness of our model, we parametrized the 8-curve such that the robot drives the right lobe backwards, and the left lobe forwards. The slight overshooting of the actual trajectory beyond the specified path is to be expected, as we only optimized for position constraints, but not for velocity constraints, when computing the control signals.

For comparison, we also trained a data-driven dynamics model using a neural network, whose input was the difference in generalized position values between the current state and the next state, and whose output was the applied control signals. It had 3 input nodes, 32 hidden nodes, and 4 output nodes. The output was normalized to lie in the interval $[-1,1]$ during training time, and then rescaled during testing.
We trained this neural network using the recorded trajectories shown in Figure~\ref{fig:ground-truth}, and used it to predict the control signals required to move between consecutive way-points. These results are shown in red in Figures~\ref{fig:path-following-circle} and~\ref{fig:path-following-eight} (also see accompanying video). The motion is very fast, and the robot quickly goes out of the field of view. We conclude that the training data is not sufficient for the neural network to learn the correct robot dynamics. In contrast, our method benefits from the data efficiency of analytical models.
\section{Conclusion and Future Work}
\label{sec:conclusion}

We presented the design of a low-cost mobile robot, as well as an analytical model that closely describes real-world recorded trajectories. To estimate unknown friction coefficients, we designed a hybrid approach that combines the data efficiency of differentiable physics engines with the flexibility of data-driven neural networks. Our proposed method is computationally efficient and more accurate than existing methods. Using our learned model, we also showed the robot drive autonomously along pre-specified paths.

In the future, we would like to improve our analytical model to also account for control signals that are not powerful enough to induce rotation of the wheels. In such cases, the wheel is not completely static and can still turn during robot motion, due to inertia. This causes the robot to change orientations in a manner that cannot be predicted by our current model. Accounting for such control signals would allow for more versatile autonomous control of our robot.

\bibliographystyle{IEEEtran}
\bibliography{bib/references.bib}

\end{document}